%% file: main.tex
\documentclass[letterpaper, 10 pt, journal]{ieeeconf}  % Comment this line out if you need a4paper

\IEEEoverridecommandlockouts                              
\usepackage{url}
\usepackage{balance}
\usepackage{amsmath}
\usepackage{amssymb,bm}
\usepackage{amsfonts}
\usepackage{enumerate}
\usepackage{tabulary}
\usepackage{multirow}
\usepackage{cite}
\usepackage{lipsum}  
\usepackage{graphicx}
\usepackage{epstopdf}
\usepackage[misc]{ifsym}
\usepackage{color}
\usepackage{xcolor}
\usepackage{xxcolor}
\usepackage{pgf}
\usepackage{pgfplots}
\usepackage{tikz}
\usepackage{float}
\usepackage{lipsum}
\usepackage{tabularx}
\usepackage{lipsum}
\usepackage{array}

\usepackage{booktabs} % For better table lines
\usepackage{array}    % For better alignment
\usepackage{graphicx} % For including images
\newcolumntype{P}[1]{>{\centering\arraybackslash}p{#1}}

\input{notation}

\title{\LARGE \bf
Radar Teach and Repeat: Architecture and Initial Field Testing}

\author{Xinyuan Qiao*, Alexander Krawciw*, Sven Lilge, and Timothy D. Barfoot% <-this % stops a space
\thanks{* Denotes equal contributions.}
\thanks{The authors are with the University of Toronto Robotics Institute, University of Toronto, Toronto, Ontario, Canada. \\ \texttt{ \{samxinyuan.qiao, alec.krawciw\}@mail.utoronto.ca}}
}

\begin{document}

\maketitle
\thispagestyle{empty}
\pagestyle{empty}

%%%%%%%%%%%%%%%%%%%%%%%%%%%%%%%%%%%%%%%%%%%%%%%%%%%%%%%%%%%%%%%%%%%%%%%%%%%%%%%%
\input{sections/0.abstract}

%%%%%%%%%%%%%%%%%%%%%%%%%%%%%%%%%%%%%%%%%%%%%%%%%%%%%%%%%%%%%%%%%%%%%%%%%%%%%%%%
% \section{LIST OF FIGURES AND TABLES}

% \begin{itemize}
%     \item \textbf{Figure 1}, Top down view: GPS path plots with a hazardous areas highlighed proving radar is robust when lidar fails (multiple paths), also include figure of Warthog with sensor (separate figure?)
%     \item box plots for pathtracking errors for both lidar and radar (many runs) normal 
%     \item box plots for pathtracking errors for both lidar and radar (many runs) challenging
%     \item radar pointcloud vs lidar pointcloud (normal vs challenging)
%     \item table: localization errors/path tracking errors
%     \item figure: system architecture overview
% \end{itemize}

\input{sections/1.intro}

\input{sections/3.methodology}

\input{sections/4.experiments}

\input{sections/5.results}

\input{sections/7.conclusion}

%\section*{APPENDIX}
%\section*{ACKNOWLEDGMENT}

\newpage
\bibliographystyle{./IEEEtranBST/IEEEtran}
\bibliography{./IEEEtranBST/IEEEabrv,./biblio}

\end{document}

%% file: notation.tex
\usepackage{xparse}

\NewDocumentCommand\bbm{}{ \begin{bmatrix} }
\NewDocumentCommand\ebm{}{ \end{bmatrix} }

\NewDocumentCommand\Matrix{m}{ \boldsymbol{\mathbf{#1}} }

\NewDocumentCommand\ArgMin{m}{ \operatorname*{argmin}_{#1} }

\NewDocumentCommand\Real{}{ \mathbb{R} }

\NewDocumentCommand\LieGroupSE{m}{ \mathrm{SE}(#1) }
\NewDocumentCommand\LieAlgebraSE{m}{ \mathfrak{se}(#1) }

\NewDocumentCommand\CoordinateFrame{m}{ \underrightarrow{\Matrix{\mathcal{F}}}_{#1} }

\NewDocumentCommand\Transform{}{ \Matrix{T} }

%% file: sections/0.abstract.tex
\begin{abstract}

Frequency-modulated continuous-wave (FMCW) scanning radar has emerged as an alternative to spinning LiDAR for state estimation on mobile robots.
Radar's longer wavelength is less affected by small particulates, providing operational advantages in challenging environments such as dust, smoke, and fog. This paper presents Radar Teach and Repeat (RT\&R): a full-stack radar system for long-term off-road robot autonomy. 
RT\&R can drive routes reliably in off-road cluttered areas without any GPS. 
We benchmark the radar system's closed-loop path-tracking performance and compare it to its {3D} LiDAR counterpart.
11.8 km of autonomous driving was completed without interventions using only radar and gyro for navigation. 
RT\&R was evaluated on different routes with progressively less structured scene geometry.
RT\&R achieved lateral path-tracking root mean squared errors (RMSE) of 5.6 cm, 7.5 cm, and 12.1 cm as the routes became more challenging. 
On the robot we used for testing, these RMSE values are less than half of the width of one tire (24 cm).
These same routes have worst-case errors of 21.7 cm, 24.0 cm, and 43.8 cm.
We conclude that radar is a viable alternative to LiDAR for long-term autonomy in challenging off-road scenarios. 
The implementation of RT\&R is open-source and available at: \url{https://github.com/utiasASRL/vtr3}.
\end{abstract}

%% file: sections/1.intro.tex
\section{INTRODUCTION}
For autonomous mobile field robots, engaging in repetitive operations within known environments, such as security patrols, building inspections, underground mining, agricultural monitoring, and campus shuttles, introduces unique challenges. 
A primary concern in these applications is effective localization when environmental artifacts may compromise visibility and access to GPS is unavailable.

Vision-based approaches to navigation are susceptible to illumination and viewpoint changes. 
Visual odometry typically fails in complete darkness \cite{liu2018towards}. 
Recent works have considered style transfer between night and day to improve feature matching at night \cite{Chen2022}.
Although LiDAR is illumination-invariant, existing methods struggle with extreme environmental artifacts such as heavy snow, dense fog, dust particles, and impenetrable smoke \cite{Harlow2023}.

Frequency-modulated continuous-wave (FMCW) radar sensors have recently demonstrated their potential for robust localization due to their longer wavelength improving resilience to environmental artifacts \cite{hong2022radarslam, Burnett2023}. 
Apart from robustness, radar can provide a significantly longer range than LiDAR \cite{Navtech}, allowing for further detection of the surrounding environment. 
This makes radar-based odometry and localization particularly effective in off-road conditions where LiDAR and cameras struggle. 
Moreover, FMCW radar maps are two-dimensional, often requiring less storage than LiDAR for large-scale maps. 
%For these reasons, radar increasingly attracts interest in the navigation literature, the recent The Oxford Offroad Radar Dataset (OORD) \cite{gadd2024oord} for example.
% The Oxford Offroad Radar Dataset (OORD) \cite{gadd2024oord} provides radar data collected in challenging mountainous terrains, further facilitating research in this domain.

\begin{figure}[t]
    \centering
    \includegraphics[width=\linewidth]{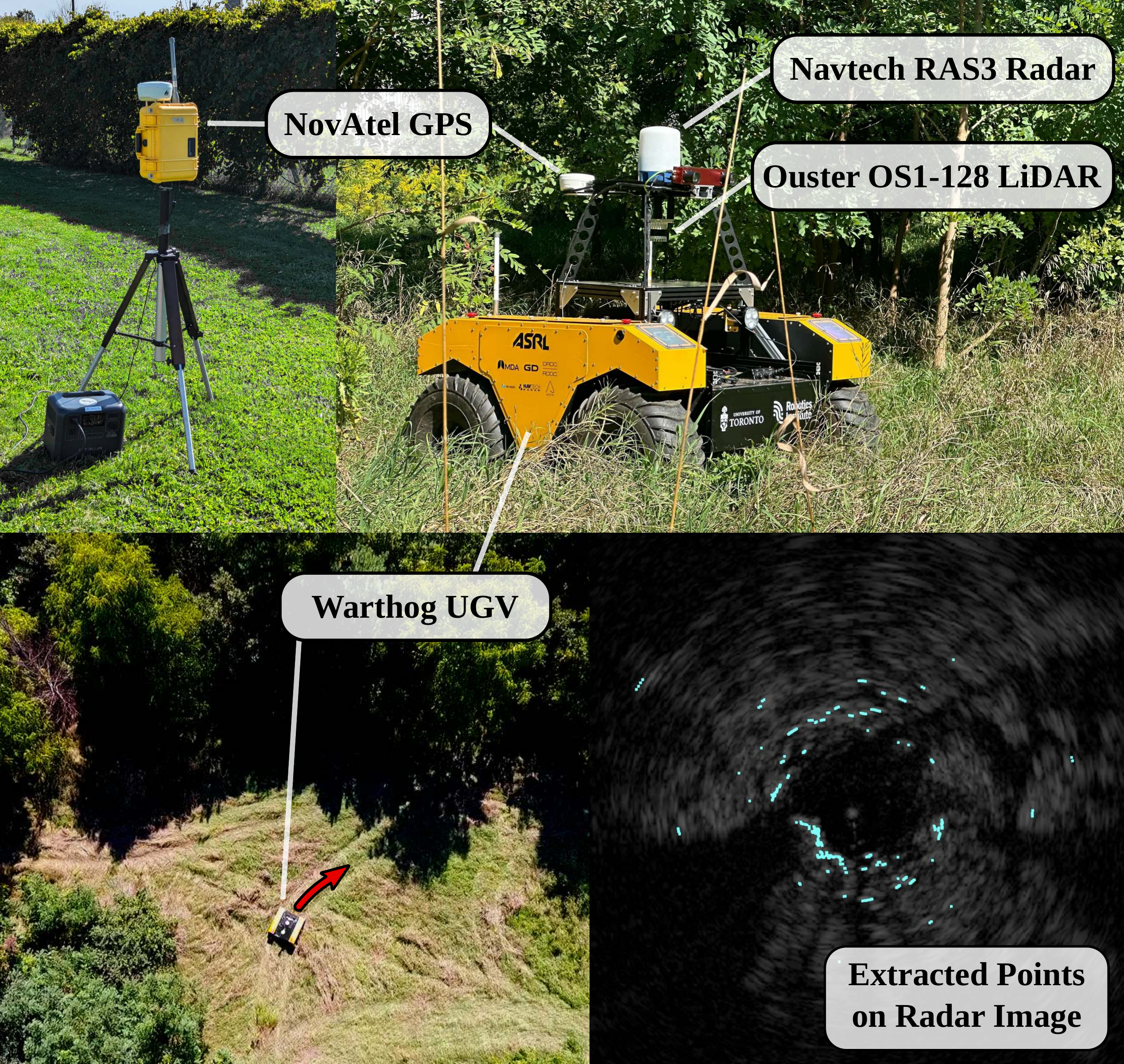}
    \caption{Top: Our Clearpath Warthog UGV is equipped with Navtech RAS3 radar, Ouster OS1-128 LiDAR, and NovAtel RTK GPS system with a stationary receiver. Bottom: Birds-eye-view of Warthog repeating a path in a clearing of the Woody Loop. The robot is repeating within the tracks, visible from the teach. 
    The associated radar point cloud highlights that Radar Teach and Repeat can drive precisely even in geometrically challenging regions.}
    \label{fig:warthogBEV}
    \vspace*{-5mm}
\end{figure}

In this paper, we extend the current Visual Teach and Repeat 3 (VT\&R3) framework\footnote{https://github.com/utiasASRL/vtr3} \cite{Furgale2010}, closing the loop for complete Radar Teach and Repeat (RT\&R).
RT\&R is implemented, tested, and evaluated onboard a Clearpath Warthog Unmanned Ground Vehicle (UGV).
Over 10 km of successful autonomous repeats were driven on and off-road using a scanning radar and a single-axis gyro. 
Figure~\ref{fig:warthogBEV} shows the Warthog driving autonomously in its tracks.
Path-tracking performance is compared with the existing LiDAR Teach and Repeat (LT\&R) functionality. 

To the best of the authors' knowledge, closed-loop driving with radar as the only exteroceptive sensor in off-road environments has not been shown before.
It therefore constitutes the primary novel contribution of this work.

\subsection{Related Work} 

\subsubsection{Teach and Repeat}
Teach and Repeat (T\&R) is a navigation paradigm with a teaching phase where local submaps are constructed through manual driving, followed by an autonomous repeat phase \cite{Furgale2010}. 
Due to its simplicity, consistency, and computational efficiency, T\&R has proven its effectiveness using sensors such as stereo vision \cite{Furgale2010} and, more recently, 3D  LiDAR \cite{Sehn2023,Burnett2022, Krusi2017} in many conditions.
The use of local topometric submaps eliminates the need for dense map construction and improves resilience to odometry drift, making T\&R robust, and well-suited for long-term robot autonomy.

\subsubsection{Why Use Radar?}

Radar is well suited to long-term autonomy for a few reasons. 
The primary advantage comes from the use of radio waves instead of visible or near-IR light found in cameras and LiDARs, respectively.
The longer wavelength means particulate matter does not interfere with the signal \cite{Cen2018}, allowing operations in extreme environments.

In contrast, LiDAR systems detect dust particles, snowflakes, and dense vegetation, which rarely correspond to permanent structures of a map \cite{Mielle2019,Ort2020,Courcelle2023}.
Corruptions to the input of the navigation pipeline can be difficult to reject in later stages. 
Another advantage of radar, shared by LiDAR, is an invariance to the ambient illumination. 
In field robotics, illumination is impossible to control and is a common failure mode of visual pipelines \cite{liu2018towards}, although there are some robust implementations of feature detectors \cite{Churchill2013, Gridseth2021, Chen2022}.

\subsubsection{Radar Odometry and Localization}
Overbye et al. \cite{overbye2023radar} use radar for mapping and terrain estimation for closed-loop navigation.
However, their framework requires accurate odometry measurements from a LiDAR scanner.
A GPU-accelerated voxel mapping system (G-VOM) \cite{overbye2022g} is adapted from LiDAR mapping to use radar for local mapping and path planning. 

To improve the behaviour of a quadrotor in fog, Michalczyk et al. \cite{michalczyk2023radar} implement radar-inertial odometry (RIO).
A dense artificial fog was created and the quadrotor was carried through it.
Visual-inertial odometry diverges almost immediately, failing to track useful features, whereas RIO remains robust and estimates the state despite the fog. 
Mielle et al. \cite{Mielle2019} compare LiDAR and radar simultaneous localization and mapping (SLAM).
They assert that radar retains essential map details, performing SLAM effectively despite lower accuracy in mapping. 

A full radar-based graph-SLAM system is developed by Hong et al. \cite{hong2020radarslam}
for reliable localization and mapping in large-scale environments with various adverse weather conditions. 
They convert FMCW radar images to point clouds and estimate radar poses online with keyframe-based pose tracking. 
Bundle adjustment is performed for local mapping and loop closures are detected using a rotation-invariant global descriptor designed for 3D point clouds. 
They show that state-of-the-art vision-based and LiDAR-based SLAM systems fail to finish some testing sequences or drift quickly in fog, snow, and heavy rain \cite{hong2020radarslam}. 

Conservative Filtering for Efficient and Accurate Radar Odometry (CFEAR) has been improving since its introduction \cite{Adolfsson2021}.
By utilizing the $k$-strongest returns per azimuth, a sparse set of oriented surface points are extracted for scan matching.
Subsequent versions, CFEAR-2 \cite{adolfsson2021cfear} and CFEAR-3 \cite{Adolfsson2022}, significantly reduce radar odometry drift by 
 addressing motion distortion, and refining filtering techniques.
CFEAR-3 adds localization to the pipeline and the method achieves performance approaching LiDAR SLAM \cite{Adolfsson2022}.

Building on CFEAR-3, Frosi et al. \cite{frosi2023advancements} propose a comprehensive radar odometry pipeline consisting of filtering,
motion compensation, oriented surface points computation, smoothing, one-to-many radar scan registration, and pose refinement.
Their work is a strong example of how radar approaches the performance of LiDAR in a pipeline. 
Importantly, Burnett et al. \cite{Burnett2022} lay the foundation for a radar-only localization pipeline using the Iterative Closest Point (ICP) method that can approach the accuracy of a LiDAR-only setup with a considerably smaller map size. 
% With high-quality odometry and localization methods, Radar Teach and Repeat has all the building blocks for a full autonomy stack. 

\subsection{Contributions}
While Burnett et al. \cite{Burnett2022} demonstrated the effectiveness of radar-only odometry and localization on the Boreas dataset \cite{Burnett2023}, a closed-loop Teach and Repeat system was not established using radar. 
Moreover, radar's performance in off-road challenging environments was not considered and tested. 
Therefore, we present a complete closed-loop Radar T\&R system and evaluate its path-tracking performance on and off-road with the existing LiDAR T\&R \cite{Burnett2022} implementation as a baseline.

The contributions of this work are
\begin{enumerate}
	\item the first closed-loop Radar Teach and Repeat pipeline,
        \item a comparison of lateral path-tracking performance with LiDAR Teach and Repeat,
        \item an open-source implementation integrated with the current VT\&R3 repository.
\end{enumerate}

The remainder of this paper is organized as follows.
Section~\ref{sec:methods} describes the methodology of this work, outlining the implementation of the RT\&R pipeline.
The experiments are discussed in Section~\ref{sec:experiments}, and Section~\ref{sec:results} summarizes the results.
Finally, Section~\ref{sec:discussion} analyzes the findings of this work and outlines the lessons learned, before Section~\ref{sec:conclusion} concludes the paper, suggesting future avenues of study.

%% file: sections/3.methodology.tex
\section{METHODOLOGY}
\label{sec:methods}
The following section provides the details of the implemented teach-and-repeat pipeline, which builds on the work of \cite{Burnett2022}.
Radar Teach and Repeat is the first closed-loop implementation of a radar-gyro autonomy system. 
Teach and Repeat consists of two stages. 
During the teach stage, a human pilot manually demonstrates a safe route to the robot. 
While driving, a sequence of submaps is constructed in a chain linked by odometry.
By mapping the environment in this way, metric localization can be performed within a submap without enforcing the global consistency of a large chain-like map. 
Once an arbitrary network of paths has been created, the robot can repeat portions of the network autonomously. 
Our system uses a 2D scanning radar, which operates at 4 Hz and a single-axis (yaw) gyro operating at 100 Hz.
The gyro improves the odometry orientation estimates significantly compared to radar ICP alone.

\subsection{Radar Scan Preprocessing and Point Cloud Extraction}
We use a customized version of the Bounded False Alarm Rate (BFAR) \cite{alhashimi2021bfar} detector to extract key points from each azimuth of the radar polar image data.
The lower-right panel of Figure~\ref{fig:warthogBEV} shows the extracted points overlaid on the raw scan image.
Similar to Cen and Newman \cite{Cen2018}, peak detection is carried out by calculating the centroid of the contiguous groups of detections.
The polar detections are converted into a 2D Cartesian point cloud for radar odometry and mapping.

\subsection{Teach Phase}

The teach phase consists of radar-gyro odometry and radar point cloud mapping construction while a human pilot manually operates the robot.

\subsubsection{Radar Continuous-Time ICP Odometry}
\begin{figure}[tbp]
    \centering
    \includegraphics[width=0.95\linewidth]{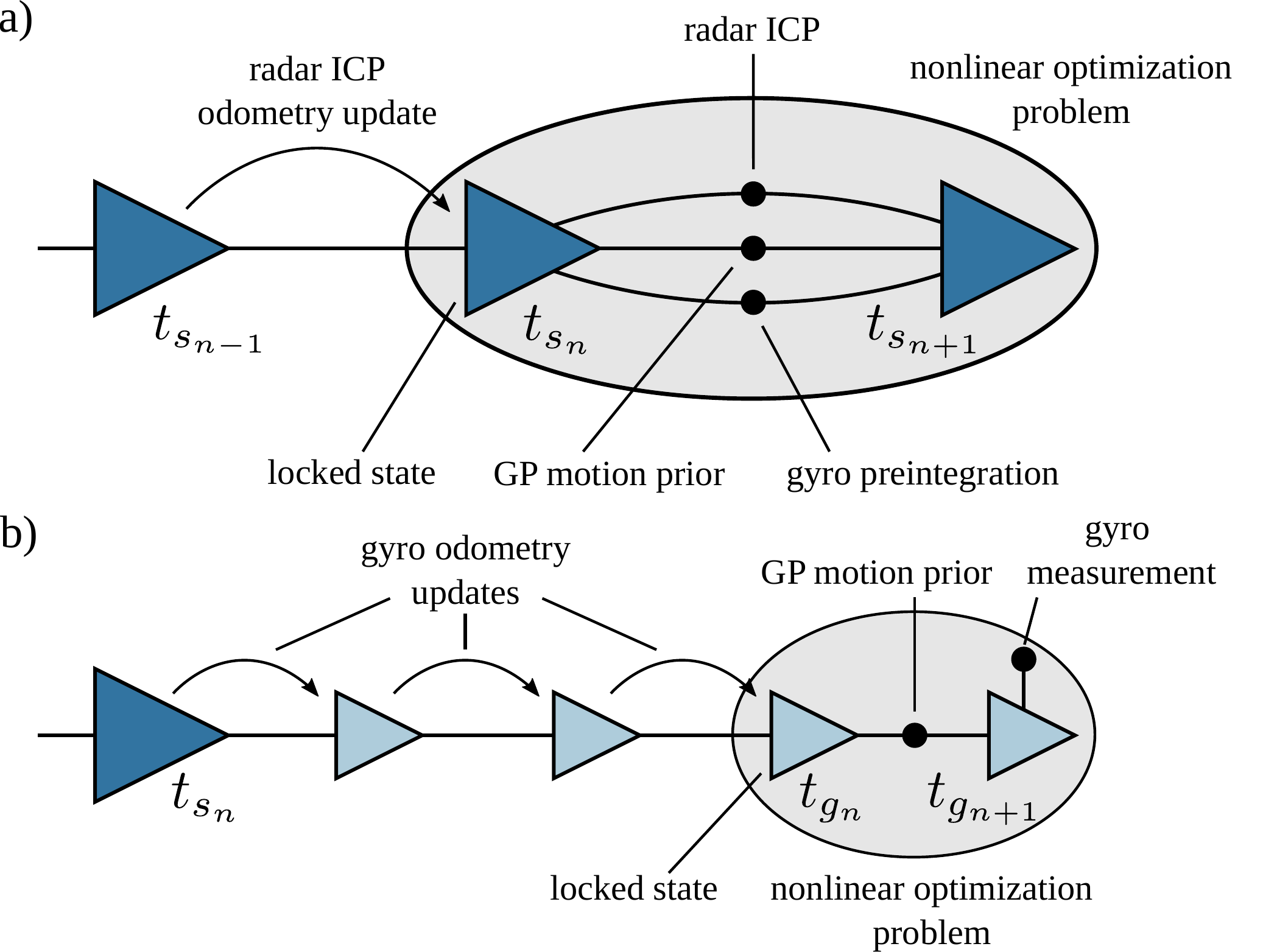}
    \vspace*{-2mm}
    \caption{A factor-graph representation of the two odometry updates. Dark blue triangles represent states containing radar scans, light blue triangles represent states containing gyro measurements, black dots represent cost terms in the optimization, and grey ellipses capture which states and cost terms are active in each optimization problem. a) The 4 Hz radar odometry pipeline uses continuous-time ICP, a constant velocity motion model, and a preintegrated yaw rate measurement term. The scan times are denoted with $t_{s_n}$ and states at which scans occur are shown as large, dark blue triangles. b) Between consecutive radar scans, the odometry is updated at 100 Hz using raw gyro yaw rate measurements. Gyro odometry updates occur between scan times $t_{s_n}$.}
    \label{fig:odometry}
    \vspace*{-4mm}
\end{figure}

A continuous-time ICP method is the backbone of our radar odometry pipeline, and we refer the reader to \cite{Burnett2022} for details on its basics without gyro factors.
%CT-ICP uses 2-D point cloud scan registration to obtain relative transformations $\Transform_{k+1,k} \in \LieGroupSE{3}$  between the current robot frame $\CoordinateFrame{r}$ and the sliding map frame $\CoordinateFrame{m}$ where we regard the whole trajectory as an exactly sparse Gaussian with Process (GP) regression problem with motion prior defined as in \cite{barfoot2024state},
%\begin{align}
%    \dot{\Transform} (t) &= \bm{\varpi}^{\wedge} \Transform (t), \\
%    \dot{\bm{\varpi}} &= \mathbf{w}(t), \mathbf{w}(t) \sim \GaussianProcess{\mathbf{0}}{\mathbf{Q}_c \delta{}(t-\tau)}.
%\end{align}
%Here, the $\wedge$ operator lifts an element of $\Real{}^6$ into a member of Lie algebra, $\LieAlgebraSE{3}$ and $\mathbf{w}(t)$ refers to a zero-mean white-noise Gaussian process.
For each scan of the radar point cloud, a nonlinear optimization problem is minimized for the state $\mathbf{x}_i = \{ {\Transform}_i, \bm{\varpi}_i \}$ of the robot, which is a pose-velocity pair. 
The timestamp of $\mathbf{x}_i$ is the timestamp of the middle azimuth of the $i$-th radar scan.
The relative transform $\Transform_{k+1,k} \in \LieGroupSE{3}$ is the estimated robot pose with respect to the last incoming radar scan at estimation node $k$. 
Odometry estimation is formulated as a nonlinear optimization with motion prior, radar measurements, and preintegration error terms derived from gyro measurements.
A Gauss-Newton solver is used to minimize the problem. 
We define the following cost function:
\begin{multline} {J}_{\text{odom,radar}} = \frac{1}{2} \mathbf{e}_{\text{prior}}^T \mathbf{Q}_\text{prior}^{-1} \mathbf{e}_{\text{prior}} + \sum_{j=1}^{M} \frac{1}{2} \mathbf{e}_{\text{odom,j}}^T \mathbf{R}_{j}^{-1} \mathbf{e}_{\text{odom,j}} \\
	 + \frac{1}{2} \mathbf{e}_{\text{yaw}}^T \mathbf{R}_\mathrm{yaw}^{-1} \mathbf{e}_{\text{yaw}}.
\end{multline}
This cost function contains three error terms as visualized in Figure~\ref{fig:odometry}a.
Here, $\mathbf{e}_\text{prior}$ is the error term of a constant-velocity motion prior with associated covariance $\mathbf{Q}_\text{prior}$ \cite{anderson2015full}.
Furthermore, \( \mathbf{R}_j \) is a constant covariance matrix associated with radar measurements (point-to-point) with the measurement error term defined as
\begin{align}
    \mathbf{e}_{\text{odom},j} &= \mathbf{D} (\mathbf{p}_{m}^{j} - \Transform(t_j)^{-1} \Transform_{rs} \mathbf{q}^j).
\end{align}
Here, \( \mathbf{q}^j \) is a homogeneous point with timestamp \( t_j \) from the \( i \)-th radar scan. 
\( \Transform_{rs} \) denotes the extrinsic calibration from the radar frame \( \CoordinateFrame{s} \) to the robot frame \( \CoordinateFrame{r} \), and \( \mathbf{T}(t_j) \) is the trajectory pose at \( t_j \), determined via Gaussian process (GP) interpolation based on the state variables \( \mathbf{x}_i^b \) and \( \mathbf{x}_i^{e} \) \cite{anderson2015full}.
Each homogeneous point,  \( \mathbf{p}_m^j \), from the sliding map is associated with \( \mathbf{q}^j \) in \( \CoordinateFrame{m} \).
Finally, \( \mathbf{D} \) removes the fourth homogeneous component.
No Doppler correction is applied as the Warthog UGV drives slowly compared to on-road vehicles in urban environments.

Lastly, a preintegration error term $\mathbf{e}_{\text{yaw}}$ with covariance $\mathbf{R}_\text{yaw}$ is included in the optimization problem.
For this error term, we integrate the yaw rate measurements of the gyro aggregated between two consecutive radar scans.
The resulting integrated term expresses an expected change in the yaw orientation of the robot pose.
The error term evaluates this expected value against the current state variables during optimization.
This concept is similar to what has been presented in \cite{Forster2015}.
Details are omitted for the sake of space.

\subsubsection{Mapping}

As the robot moves along the teach path, a window of select point clouds is merged and stored as a local submap. 
A new submap is created if the robot has changed position or orientation by more than a specified threshold. 
The relative odometry transformation between each submap is used to decide against which submap to localize during a repeat. 
The graph of these relative poses and associated data forms the basis of a pose graph.

\subsection{Repeat Phase}
Once the robot transitions to driving autonomously, localization is added to the processing pipeline.
The same 4 Hz radar-gyro odometry is run during the repeat.

\subsubsection{High-Rate Gyro Odometry} 
The 4 Hz update rate of the radar sensor was observed to be too slow for precise path tracking in RT\&R. 
To provide more frequent odometry estimates, another small optimization problem is solved for each incoming gyro measurement at 100 Hz that uses the current gyro measurement as well as the last estimated robot state.
The cost function for this optimization problem is
\begin{multline}
	{J}_{\text{odom,gyro}} = \frac{1}{2} \mathbf{e}_{\text{prior}}^T \mathbf{Q}_\text{prior}^{-1} \mathbf{e}_{\text{prior}} + \frac{1}{2} \mathbf{e}_{\text{gyro}}^T \mathbf{R}_\mathrm{gyro}^{-1} \mathbf{e}_{\text{gyro}},
\end{multline}
where $\mathbf{e}_\text{prior}$ and $\mathbf{Q}_\text{prior}$ are again the error term and covariance of the motion prior \cite{anderson2015full}, and $\mathbf{e}_{\text{gyro}}$ and $\mathbf{R}_\mathrm{gyro}$ are the error term and covariance for the current yaw rate measurement (see Figure~\ref{fig:odometry}b).
This effectively allows for a high-rate extrapolation of the motion model refined by yaw rate measurements between the lower-rate radar scans. 
High-rate odometry results are replaced with the more accurate ICP odometry results, as soon as a new radar scan is available.

\subsubsection{Localization ICP}
Between the live radar scan and a stored submap point cloud, ICP is used to obtain the relative transformation $\hat{\Transform}_{rl}$ connecting the robot frame $\CoordinateFrame{r}$ to the local map of the spatially closest vertex frame $\CoordinateFrame{l}$ of the teach path. 
A nonlinear localization cost function,
\begin{align}
    {J}_\text{loc} &= \frac{1}{2} \mathbf{e}_{\text{pose}}^T \mathbf{Q}_\text{pose}^{-1} \mathbf{e}_{\text{pose}} + \sum_{j=1}^{M}\frac{1}{2}\mathbf{e}_{\text{loc}, j}^{{T}} \mathbf{R}_{j}^{-1} \mathbf{e}_{\text{loc},j}, \\ \mathbf{e}_{\text{loc},j} &= \mathbf{D}(\mathbf{p}_{l}^{j} - \Transform_{rl}^{-1} \Transform_{rs} \mathbf{q}^j),
\end{align}
consists of a pose prior based on the latest odometry estimate, and the point-to-point ICP cost terms. 
Additionally, $\mathbf{q}^j$ is a live-scan homogeneous point and the nearest local map point to $\mathbf{q}^j$ expressed in the localization vertex frame $\CoordinateFrame{l}$ is $\mathbf{p}_{l}^{j}$. 
Furthermore, $\mathbf{e}_{\text{pose}}$ and $\mathbf{Q}_{\text{pose}}$ are the error term and covariance of a prior pose, which is obtained using the current odometry estimate and the last solved localization against the teach path.

\subsubsection{Closed-Loop Control}
The solution of the optimization problem is passed to a model-predictive controller (MPC) \cite{Rawlings2022} for accurate path tracking.
The control input to the robot $\mathbf{u}_k$ contains a linear ($v_k$) and angular ($\omega_k$) velocity command. 
The low-level controller is responsible for tracking those commands. 
A unicycle motion model with a first-order lag,
\begin{equation}
\label{eq_laggy_mm}
    \dot{f}(\Transform_{k-1}, \mathbf{u}_k, \mathbf{u}_{k-1}) = \begin{bmatrix}
        \dot{x} \\
        \dot{y} \\
        \dot{\theta}
    \end{bmatrix} = 
    \begin{bmatrix}
        v_k  \cos \theta_{k-1} \\
        v_k \sin \theta_{k-1} \\
        (1 - \alpha) \omega_k + \alpha \omega_{k-1}
    \end{bmatrix}
\end{equation}
is optimized over a finite horizon of $K$ target poses along the teach path.
The constant $\alpha$ represents the time response of the system and was tuned to 0.4 for the Warthog. 
Although three-dimensional state estimation is performed, the wheeled robot can only be controlled in the plane. 
The nearby curvature is captured as precisely as possible using the robot's local plane. 
For each reference pose $\Transform_{\text{ref},k} \in \LieGroupSE{2}$ there is an associated rolled-out pose 
$\Transform_{k} \in \LieGroupSE{2}$.
Furthermore, $\Transform_0$ and $\mathbf{u}_0$ define the current state of the robot.
The $\vee$ operator transforms elements of $\LieGroupSE{2}$ into the $\Real{}^3$ vector in the Lie Algebra $\LieAlgebraSE{2}$.
Path tracking is achieved by interpolating along the curvature of the reference path to set the reference poses for each optimization. 
The MPC is a constrained nonlinear optimization,
\begin{subequations}
\label{eq:mpc}
\begin{multline}
     \ArgMin{\Transform, \mathbf{u}} \sum_{k=1}^K \ln \left(\Transform_{\text{ref},k} \Transform_k^{-1} \right)^{\vee^{T}} \mathbf{Q}^{-1} \ln \left(\Transform_{\text{ref},k} \Transform_k^{-1} \right)^{\vee} \\ + \mathbf{u}_k^{T} \mathbf{R}^{-1} \mathbf{u}_k
\end{multline}
subject to $ \forall \  k \in 1, 2, ..., K$
\begin{align}  
\label{eq:mm_con}
    &\Transform_{k} = f(\Transform_{k-1}, \mathbf{u}_k, \mathbf{u}_{k-1}) \\
\label{eq:vel_con}
    &\mathbf{u}_{\text{min}} < \mathbf{u}_k < \mathbf{u}_\text{max} \\
\label{eq:corr_con}
    &c_\text{right} < (\mathbf{p}_k^b - \mathbf{p}_k^a)^{{T}} (\mathbf{D}\Transform_{k} - \mathbf{p}_k^a) < c_{\text{left}},
\end{align}
\end{subequations}
solved using CasADi \cite{Andersson2019}.
The weighting terms $\mathbf{Q}$ and $\mathbf{R}$ are constant matrices for the pose error and velocity penalties, respectively. 
The motion-model equality constraint \eqref{eq:mm_con} uses a fourth-order Runge-Kutta numerical integration of the nonlinear motion model \eqref{eq_laggy_mm}.
The time step between poses is $0.25$ s to match the radar's scan rate.
Constraints are imposed on the solution space to ensure the safety of the vehicle. 
The velocity is constrained \eqref{eq:vel_con} with constant maximum forward and reverse speeds. 
The corridor constraint \eqref{eq:corr_con} defines the allowable lateral path deviations $c_\text{right}$ and $c_\text{left}$ around the robot.
In our experiments, the maximum lateral deviation is set to 50 cm. 
$\mathbf{D}$ is the projection matrix to pick off the translation components of the transformation, and $\mathbf{p}_k^a$ and $\mathbf{p}_k^b$ are the endpoints of the closest section of the discrete teach path associated with $\Transform_{\text{ref},k}$.
If the robot violates the corridor constraint, i.e., there is no feasible solution to \eqref{eq:mpc}, the robot will halt and wait for pilot intervention. 

% \lipsum[1-2]

%% file: sections/4.experiments.tex
\section{EXPERIMENTS}
\label{sec:experiments}
To evaluate Radar Teach and Repeat, a progression of structured to unstructured environments was selected to demonstrate that robot performance is maintained if the scene degrades. 
Lateral path-tracking error is the primary metric and is evaluated by comparing real-time kinematic (RTK) GPS data from the teach and repeat phases. 
In total 20 km of repeats were completed, of which 11.8 km were RT\&R.

\subsection{Clearpath Warthog Hardware Platform}
A Clearpath Warthog  UGV \cite{Warthog}, shown in Figure~\ref{fig:warthogBEV}, is equipped with a Navtech RAS3 radar \cite{Navtech} mounted on the top plate and an Ouster OS1-128 LiDAR \cite{ouster_128} underneath. 
A NovAtel SMART6 RTK-GPS system \cite{novatelGPS} provides accurate ground truth for evaluating path-tracking performance. A stationary receiver serves as the live RTK reference.

% \begin{table}[htbp]
%     \centering
%      \caption{Sensor Suite of the Clearpath Warthog \tim{can potentially cut}}
%     \renewcommand{\arraystretch}{1.75} % Increase row height
%     \setlength{\tabcolsep}{15pt}      % Increase column separation
%     \begin{tabularx}{0.85\linewidth}{@{}>{\bfseries}l@{\hspace{15pt}} p{4cm}@{}}
%         \toprule
%         Sensor & \textbf{Specifications} \\
%         \midrule
%         Navtech RAS3 Radar & \renewcommand{\arraystretch}{1}
%         \begin{tabular}[t]{@{\textbullet~}p{3cm}@{}}
%             0.0438m range resolution\\
%             75m range\\
%             4Hz update rate\\
%             360° HFOV 
%        \end{tabular} \\
%         Ouster OS1-128 LiDAR& \renewcommand{\arraystretch}{1}
%         \begin{tabular}[t]{@{\textbullet~}p{3.5cm}@{}}
%             128 beams \\
%             1 mm range resolution\\
%             90m range\\
%             10 Hz update rate\\
%             360° HFOV x 42.4° VFOV
%        \end{tabular} \\
%         Invensense ICM-20948 Gyro &\renewcommand{\arraystretch}{1}
%         \begin{tabular}[t]{@{\textbullet~}p{3cm}@{}}
%             100 Hz update rate \\
%             $\pm250$°/sec saturation \\
%             131 LSB/°/sec accuracy 
%        \end{tabular} \\
%         Novatel SMART6 GPS & \renewcommand{\arraystretch}{1}
%         \begin{tabular}[t]{@{\textbullet~}p{3.5cm}@{}}
%             dual-frequency RTK \\
%             0.01 m +1 ppm hor. accuracy
%        \end{tabular} \\
%         \bottomrule
%     \end{tabularx}
%     \label{tab:sensor_specs}
% \end{table}

\subsection{Evaluation Routes}

\begin{figure}[b]
    \centering
    \vspace*{-4mm}
    \minipage{0.48\textwidth}
        \includegraphics[width=\linewidth]{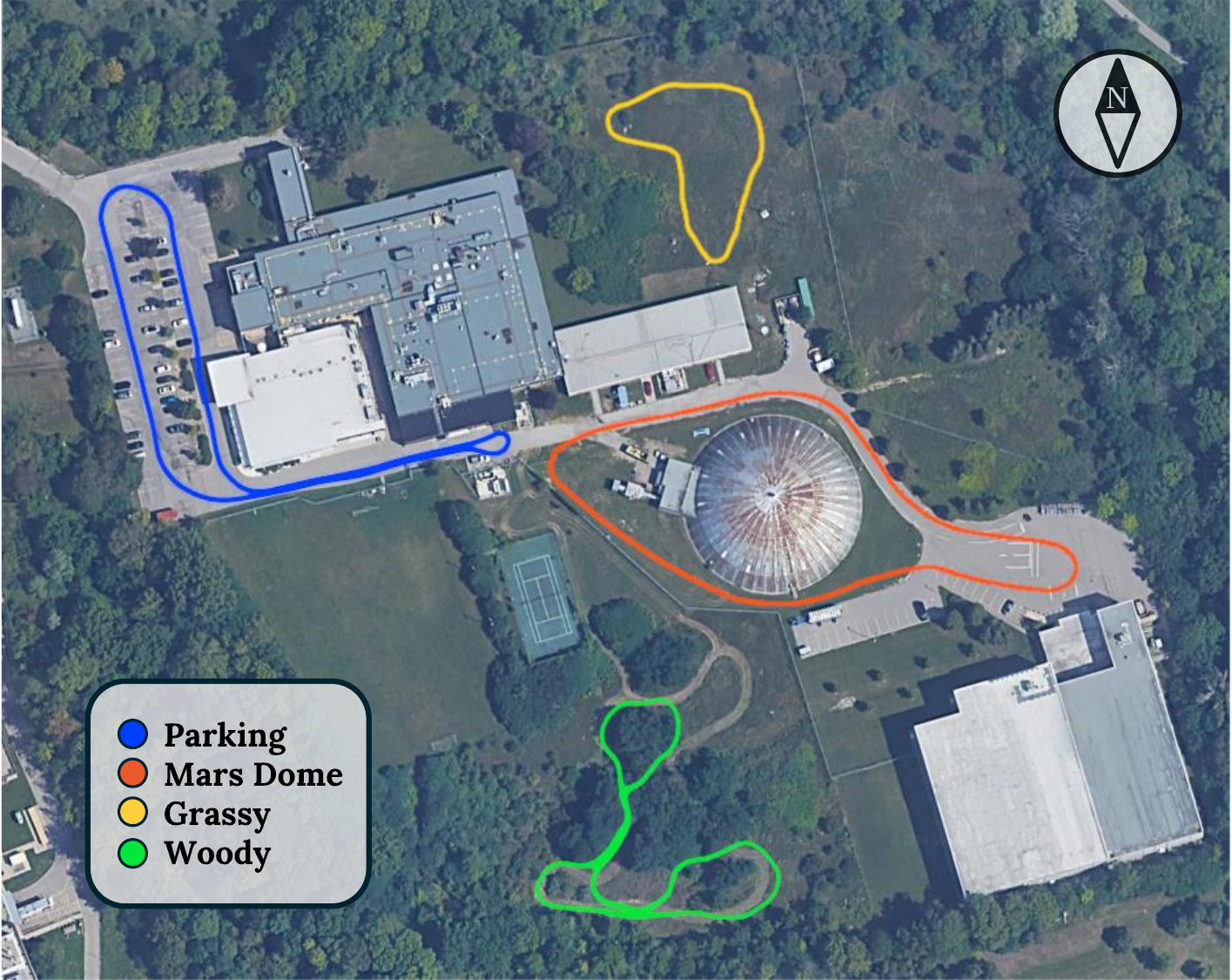}
        \endminipage\hfill
    \caption{Satellite view of the four routes used for evaluation.}
    \label{fig:routes}
\end{figure}

Radar T\&R was evaluated on four routes at the University of Toronto Institute for Aerospace Studies.
The routes consist of the Parking Loop (368 m), Mars Dome Loop (360 m), the Grassy Loop (175 m), and the Woody Loop (373 m). 
The routes are visualized in Figure~\ref{fig:routes}.
The Parking Loop and Mars Dome Loop contain a mixture of urban structures (fences and vehicles) and vegetation (trees and bushes). The area east of the Mars Dome was particularly challenging due to sparse point clouds.
The Grassy Loop is a field with elevation changes, ubiquitous potholes, and grass nearly as tall as the robot. 
A fence exists on the east side of the loop.
The Woody Loop is entirely off-road with dense vegetation and no significant geometric structure.
Parts of the route have less than 30 cm of lateral clearance and tight turns that must be traversed precisely to avoid collisions.  
Table~\ref{tab:route_info} summarizes the details of the four routes driven.

% % the combined table hahaha
% \begin{table}[t]\renewcommand{\arraystretch}{1.1}
%     \centering
%     \caption{Experimental routes completed and evaluated.}
%         \centering
%         % First table
%         \begin{tabular}{|c|>{\centering}m{0.9cm}|>{\centering}m{1.1cm}|>{\centering}m{1.1cm}|>{\centering}m{1cm}|c|}
%             \hline
%              \textbf{Path}  & \textbf{Teach (m)} &  \textbf{\# LT\&R Repeats} &  \textbf{\# RT\&R Repeats} & \textbf{Radar Dist (km)} & \textbf{LiDAR Dist (km)}\\
%             \hline \hline
%              Parking & 368 &  5 &  10 & 1.84 & 3.68\\
%             \hline
%             Mars Dome & 360 &  5 &  5 & 1.80 & 1.80 \\
%             \hline
%             Grassy & 175 &  15 &  15 &  2.63 & 2.63 \\
%             \hline
%             Woody & 373 &  5 &  10 & 1.87 & 3.73 \\
%             \hline \hline
%             \textbf{Total} & 1276 &  30 &  40 & 8.12 & 11.84 \\
%             \hline
%         \end{tabular}
%         \label{tab:route_info}
%     \vspace{-4mm}
% \end{table}

\begin{table}[t]
    \renewcommand{\arraystretch}{1.25}
    \centering
    \caption{Experimental routes completed and evaluated.}
    \begin{tabular}{|>{\centering\arraybackslash}p{1.4cm}|>{\centering\arraybackslash}m{1cm}|>{\centering\arraybackslash}m{0.3cm}>{\centering\arraybackslash}m{1.2cm}|>{\centering\arraybackslash}m{0.3cm}>{\centering\arraybackslash}m{1.2cm}|}
      \hline
      \rule{0pt}{13pt}\textbf{Path} & \textbf{Teach} &  \multicolumn{2}{c|}{\parbox{1.85cm}{\textbf{\# and Distance LT\&R Repeats}}} &  \multicolumn{2}{c|}{\parbox{1.85cm}{\textbf{\#  and Distance  RT\&R Repeats}}} \\[6pt] \hline \hline
        Parking & 368 m&  5 & (1.8 km) &  10 & (3.7 km) \\
        \hline
        Mars Dome & 360 m&  5 & (1.8 km) &  5 & (1.8 km)  \\
        \hline
        Grassy & 175 m&  15 & (2.6 km) &  15 & (2.6 km) \\
        \hline
        Woody & 373 m&  5 & (1.9 km) &  10 & (3.7 km) \\
        \hline \hline
        \textbf{Total} & 1276 m&  30 & (8.1 km) &  40 & (11.8 km)  \\
        \hline
    \end{tabular}
    \label{tab:route_info}
    \vspace{-4mm}
\end{table}

\begin{table*}\renewcommand{\arraystretch}{1.25}
 	\centering
		% Second table
		\caption{Measured and estimated path-tracking errors for LiDAR and Radar T\&R across different paths. \\ * Dense tree canopies over the Woody Loop prevented the calculation of this metric.}
  \label{tab:primaryResults}
		\begin{tabular}{|c|c|>{\centering}m{2.4cm}|>{\centering}m{2.4cm}|>{\centering}m{2.5cm}|>{\centering}m{2.6cm}|}
			\hline  
			\textbf{Sensor Modality} & \textbf{Path} & \centering \textbf{GPS-Measured Lateral RMSE (m)} & \textbf{T\&R-Estimated Lateral RMSE (m)} & \textbf{GPS-Measured Max Lateral Error (m)} & \textbf{T\&R-Estimated Max Lateral Error (m)}  \tabularnewline \hline \hline
   \multirow{4}{*}{LiDAR T\&R}  
                & Parking & 0.035 & 0.033 & 0.190 & 0.168 \tabularnewline \cline{2-6}
                & Mars Dome & 0.026 & 0.027 & 0.148 & 0.169\tabularnewline \cline{2-6}
                & Grassy  & 0.047 &  0.050 &  0.165 &  0.206 \tabularnewline \cline{2-6}
			& Woody & \textasteriskcentered & 0.043 & \textasteriskcentered & 0.147 \tabularnewline \hline \hline
           \multirow{4}{*}{Radar T\&R}  
			& Parking  & 0.056 & 0.025 & 0.217 & 0.158\tabularnewline \cline{2-6}
                & Mars Dome & 0.075 & 0.027 & 0.240 & 0.212\tabularnewline \cline{2-6}
                & Grassy  & 0.121 & 0.036 & 0.438 & 0.164 \tabularnewline \cline{2-6}			
                & Woody & \textasteriskcentered & 0.037 & \textasteriskcentered & 0.134 \tabularnewline \hline
		\end{tabular}
\end{table*}

% \begin{table*}[hb]
%     \centering
%     \caption{Details of the routes used for testing. In total over ten autonomous kilometres were driven with no interventions.}
%     \begin{tabular}{|c|c|c|c|c|}
%          \hline
%          Name & Length (m) & \# of RT\&R Repeats & \# of LT\&R Repeats & Total (km) \\
%          \hline
%          Mars Dome & 360 & 5 & 5 & 3.6 \\
%          \hline
%           Parking & 800 & 1 & 0 & 0.8\\
%          \hline
%          Woody & 280 & 1 & 0 & 0.28 \\
%          \hline
%          \textbf{Total} & 1100 & 7 & 5 & 4.68 \\
%          \hline
%     \end{tabular}
%     \label{tab:route_info}
% \end{table*}

% \subsubsection{Mars Dome Loop Route}

% \subsubsection{Parking Loop Route}

% \subsubsection{Shrubby Loop Route}

% \lipsum[1-2]

%% file: sections/5.results.tex
\section{Results}
\label{sec:results}
Qualitatively, after a single teach the robot continues to localize and follow the route reliably during repeat.
Precision maneuvers, such as parking in the garage, or passing through open gates can be performed consistently.

To evaluate the quality of Radar Teach and Repeat, the robot's GPS position during the teach and repeat is compared. 
Both radar and LiDAR sensor data were logged simultaneously during the human pilot demonstration to ensure that the radar and LiDAR teach paths were identical. 
These sensor data were used to construct the maps. 
In all repeats, only the sensor under test was activated. 

\subsection{Radar Teach and Repeat Path-Tracking Performance}

In Teach and Repeat, high-quality path-tracking means that the lateral errors are small.
The lateral path-tracking error is the perpendicular distance between the robot and the closest discrete teach-path segment. 
Our primary metric for path-tracking is the lateral root mean squared error (RMSE). 
The maximum lateral error is reported to demonstrate the worst-case combination of localization and control over all repeats.

In Table~\ref{tab:primaryResults}, the GPS-measured RMSE refers to the lateral RMSE obtained by comparing the GPS data of the teach and all repeats.
The T\&R-estimated RMSE uses T\&R's internal estimate of the lateral error based on the result of localization.
A small difference between the measured and estimated values suggests accurate localization.

RT\&R had a 100\% autonomy rate in all four environments. 
The four routes contain progressively less structure and progressively more difficult terrain for the controller. 
To highlight the capability of RT\&R to operate in GPS-denied environments, the Woody Loop is driven primarily under the tree canopy, which disrupts GPS signals, preventing a quantitative performance assessment.

Table~\ref{tab:primaryResults} has GPS-measured lateral RMSE values that increase from 5.6 cm in the Parking Loop to 12.1 cm in the Grassy Loop.
Qualitatively, the Woody Loop performance is similar to the Grassy Loop, although the GPS-measured RMSE cannot be computed. 
The supplementary video provides sample drone and first-person footage of this route being driven autonomously. 

Interestingly, the T\&R-estimated lateral RMSE is much smaller than the actual GPS-measured RMSE for the radar, ranging from 2.5 cm in the Parking Loop to 3.7 cm in the Woody Loop. 
The difference between these two values implies that RT\&R localization believes the robot is nearly on the path even if the robot is off the path. 
Incorrect lateral position estimates cause the controller to react less aggressively, leading to wide turns. This behaviour was observed more commonly in the Grassy and Woody Loops.

\begin{figure*}[ht]
	\centering
	\includegraphics[width=\linewidth]{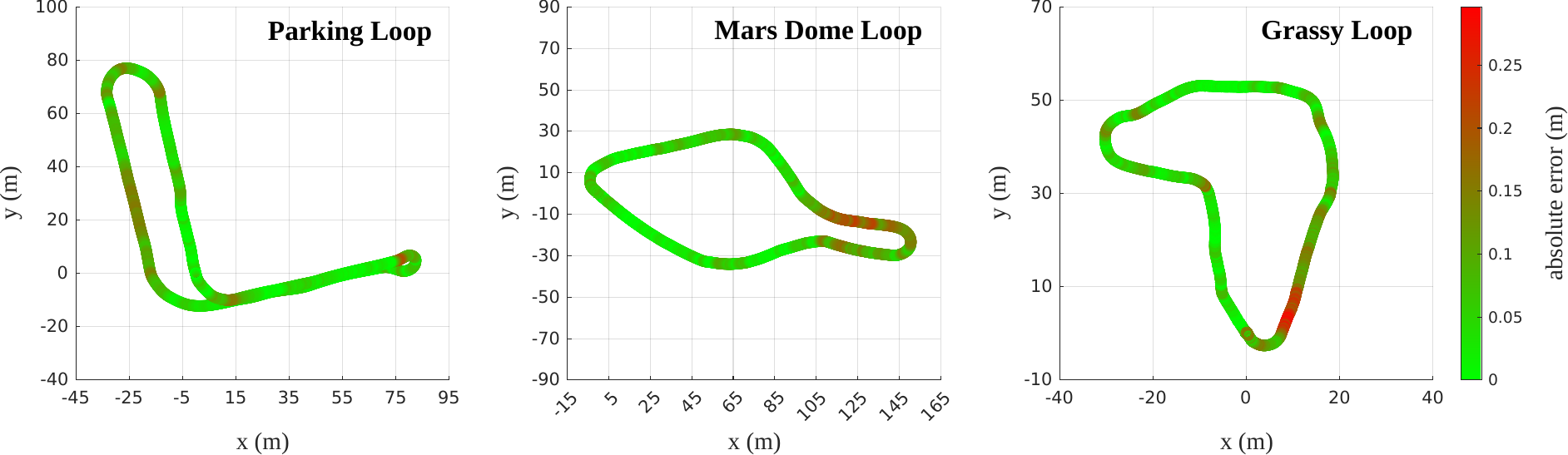}
 \vspace*{-7mm}
	\caption{Example repeats of the Parking, Mars Dome, and Grassy Loops using the RT\&R pipeline. The absolute lateral path-tracking error along these repeats is highlighted using colours ranging from green (small error) to red (large error).}
	\label{fig:path_error}
 \vspace*{-2mm}
\end{figure*}

Figure~\ref{fig:path_error} highlights the lateral error along example repeats of the Parking, Mars Dome, and Grassy Loops. 
On certain paths, the maximum errors arise from the controller rather than localization.
In the Parking Loop, the maximum error occurs after a sharp turn near the right edge of the path for both radar and LiDAR.
The consistency suggests that the shared controller is responsible.
In the Grassy Loop, the maximum error occurs in a relatively straight section of the path. 
Here, localization has a large error due to the sparse geometry.
The LiDAR and radar maximum errors on this route occur in different locations. 
This further implies that the lateral error results from localization, not control.

\subsection{Comparison to LiDAR Teach and Repeat Baseline}
To compare RT\&R to LT\&R, the same loops were repeated in closed-loop using identical controller parameters. 
LT\&R outperformed RT\&R on all four routes. 
The baseline LT\&R measurements in Table~\ref{tab:primaryResults} do not vary as radar does with environmental structure. 
The smallest GPS-measured lateral RMSE is the Mars Loop with 2.6 cm and the largest is the Grassy Loop with 4.7 cm.
Notably, when comparing the measured and estimated path-tracking errors in Table~\ref{tab:primaryResults} they are less than 1 cm different.
This suggests that LiDAR lateral path-tracking error is dominated by the controller error, not localization. 
In the Parking Loop, RT\&R performance approaches LT\&R for RMSE (5.6 cm vs 3.5 cm) and maximum lateral error (21.7 cm vs 19.0 cm). 

\begin{figure}[h]
    \centering
    \minipage{0.48\textwidth}
        \includegraphics[width=\linewidth]{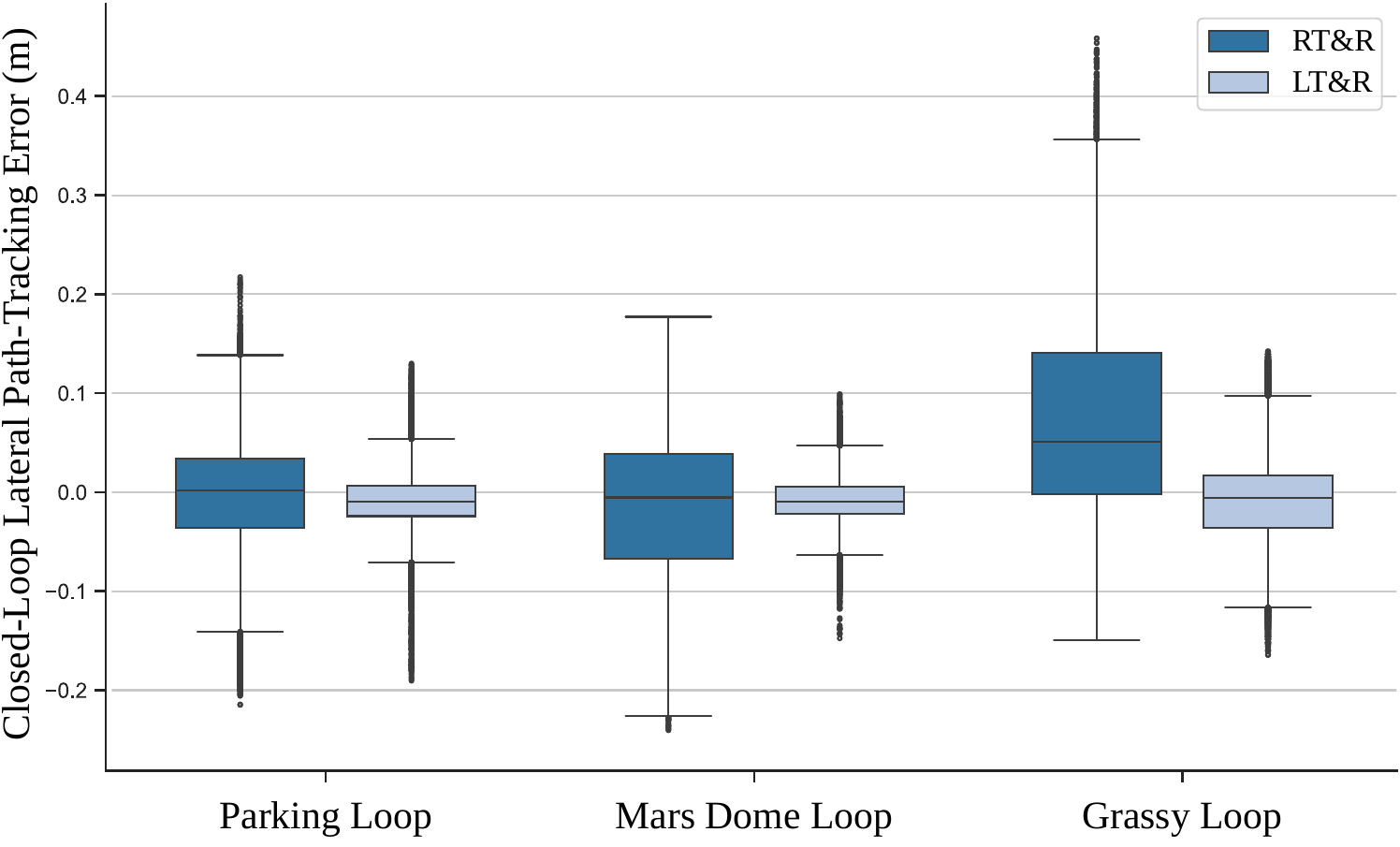}
        \endminipage\hfill
    \caption{Distribution of RT\&R and LT\&R lateral path-tracking errors across all repeats of three routes.}
    \label{fig:boxplot}
    \vspace{-5mm}
\end{figure}

Figure~\ref{fig:boxplot} provides a box plot of the raw path-tracking errors over all repeats on the Parking, Mars Dome, and Grassy Loops for both sensors. 
The spread of LiDAR errors is smaller than radar in all three cases. 
The most interesting result is the positive error bias in the Grassy Loop for RT\&R. 
This occurs because the Warthog drives outside of the curvature of the teach path. 
Localization believes that the robot is less than 5 cm from the path at the region with the maximum error so the controller takes no corrective action.
Poor localization may be attributed to pitching effects from potholes, which affect the 2D radar more than the 3D LiDAR.

\section{Discussion and Lessons Learned}

\label{sec:discussion}
Implementing closed-loop control using the FMCW radar as the only rich sensor proved challenging in multiple respects. 
A primary observation was the difficulty in filtering radial artifacts produced by the radar scans. 
These sparse rings of points are centred around the sensor and often caused ICP to favour solutions that aligned the radial artifacts especially when there was less geometry in the scene. 
This error caused heading estimates from odometry to drift rapidly.
Interestingly, heading errors became impactful because the incorrect yaw rate created a bad feedback loop with the controller resulting in the robot diverging from the path. 
Including the gyro's direct measurement of the yaw rate was crucial to overcoming these errors. 

An alternative approach to improving odometry was evaluating different point cloud extractors and tuning parameters. 
It was difficult to quantify the impact of the point-extractor tuning on the closed-loop metrics.
This suggests that there remains room for improvement in the extraction process to capture the scene with fewer false-positive points.

%% file: sections/7.conclusion.tex
\section{Conclusion}
\label{sec:conclusion}
In this work, we presented Radar Teach and Repeat: a full autonomy stack capable of driving in unstructured environments using a 2D scanning radar and a single-axis gyro. 
The continuous-time state estimation allows for seamless sensor fusion of the radar and gyro operating at different rates. 
The high-rate odometry improves the ability of RT\&R to control the Warthog UGV.
The sensor combination remains robust even in environments where the radar scans produce sparse point clouds. 
With this added robustness, we compared the closed-loop path-tracking performance of radar to LiDAR on three different routes and completed 11.8 km of intervention-free autonomous driving. 
Radar performs well in the routes tested (5.6 cm, 7.5 cm, and 12.1 cm RMSE) but future work remains to close the gap with LiDAR. 
For example, radial artifacts observed in the radar point clouds extracted cause angular drift in odometry.
Improved point-feature extraction approaches will be explored to reduce this effect.
Finally, we intend to add Doppler correction to the odometry, similar to Lisus et al. \cite{Lisus2024}. 
Improvements such as these will enhance Radar Teach and Repeat positioning it as a robust solution for long-term autonomy in off-road challenging environments.

\section{Acknowledgements}
This work was supported by the Natural Science and Engineering Research Council (NSERC). 
We thank Hexagon NovAtel, Navtech Radar, and Clearpath Robotics for their support.

%% file: main.bbl
\begin{thebibliography}{10}
\providecommand{\url}[1]{#1}
\csname url@rmstyle\endcsname
\providecommand{\newblock}{\relax}
\providecommand{\bibinfo}[2]{#2}
\providecommand\BIBentrySTDinterwordspacing{\spaceskip=0pt\relax}
\providecommand\BIBentryALTinterwordstretchfactor{4}
\providecommand\BIBentryALTinterwordspacing{\spaceskip=\fontdimen2\font plus
\BIBentryALTinterwordstretchfactor\fontdimen3\font minus \fontdimen4\font\relax}
\providecommand\BIBforeignlanguage[2]{{%
\expandafter\ifx\csname l@#1\endcsname\relax
\typeout{** WARNING: IEEEtran.bst: No hyphenation pattern has been}%
\typeout{** loaded for the language `#1'. Using the pattern for}%
\typeout{** the default language instead.}%
\else
\language=\csname l@#1\endcsname
\fi
#2}}

\bibitem{liu2018towards}
P.~Liu, M.~Geppert, L.~Heng, T.~Sattler, A.~Geiger, and M.~Pollefeys, ``Towards robust visual odometry with a multi-camera system,'' in \emph{IEEE/RSJ International Conference on Intelligent Robots and Systems}, 2018, pp. 1154--1161.

\bibitem{Chen2022}
Y.~Chen and T.~D. Barfoot, ``Self-supervised feature learning for long-term metric visual localization,'' \emph{IEEE Robotics and Automation Letters}, vol.~8, no.~2, pp. 472--479, 2022.

\bibitem{Harlow2023}
K.~Harlow, H.~Jang, T.~D. Barfoot, A.~Kim, and C.~Heckman, ``A new wave in robotics: Survey on recent mmwave radar applications in robotics,'' \emph{arXiv preprint arXiv:2305.01135}, 2023.

\bibitem{hong2022radarslam}
Z.~Hong, Y.~Petillot, A.~Wallace, and S.~Wang, ``Radarslam: A robust simultaneous localization and mapping system for all weather conditions,'' \emph{The International Journal of Robotics Research}, vol.~41, no.~5, pp. 519--542, 2022.

\bibitem{Burnett2023}
K.~Burnett, D.~J. Yoon, Y.~Wu, A.~Z. Li, H.~Zhang, S.~Lu, J.~Qian, W.-K. Tseng, A.~Lambert, K.~Y. Leung, \emph{et~al.}, ``Boreas: A multi-season autonomous driving dataset,'' \emph{The International Journal of Robotics Research}, vol.~42, no. 1-2, pp. 33--42, 2023.

\bibitem{Navtech}
\BIBentryALTinterwordspacing
{Navtech RAS3 Radar}. (2023). [Online]. Available: \url{https://navtechradar.atlassian.net/wiki/spaces/IA/pages/2573172737/RAS3}
\BIBentrySTDinterwordspacing

\bibitem{Furgale2010}
P.~Furgale and T.~D. Barfoot, ``Visual teach and repeat for long-range rover autonomy,'' \emph{Journal of Field Robotics}, vol.~27, no.~5, pp. 534--560, 2010.

\bibitem{Sehn2023}
J.~Sehn, Y.~Wu, and T.~D. Barfoot, ``Along similar lines: Local obstacle avoidance for long-term autonomous path following,'' in \emph{Conference on Robots and Vision}, 2023, pp. 81--88.

\bibitem{Burnett2022}
K.~Burnett, Y.~Wu, D.~J. Yoon, A.~P. Schoellig, and T.~D. Barfoot, ``Are we ready for radar to replace lidar in all-weather mapping and localization?'' \emph{IEEE Robotics and Automation Letters}, vol.~7, no.~4, pp. 10\,328--10\,335, 2022.

\bibitem{Krusi2017}
\BIBentryALTinterwordspacing
P.~Krüsi, P.~Furgale, M.~Bosse, and R.~Siegwart, ``Driving on point clouds: Motion planning, trajectory optimization, and terrain assessment in generic nonplanar environments,'' vol.~34, no.~5, pp. 940--984, 2017. [Online]. Available: \url{https://onlinelibrary.wiley.com/doi/abs/10.1002/rob.21700}
\BIBentrySTDinterwordspacing

\bibitem{Cen2018}
S.~H. Cen and P.~Newman, ``Precise ego-motion estimation with millimeter-wave radar under diverse and challenging conditions,'' in \emph{IEEE International Conference on Robotics and Automation}, 2018, pp. 6045--6052.

\bibitem{Mielle2019}
M.~Mielle, M.~Magnusson, and A.~J. Lilienthal, ``A comparative analysis of radar and lidar sensing for localization and mapping,'' in \emph{European Conference on Mobile Robots}, 2019, pp. 1--6.

\bibitem{Ort2020}
T.~Ort, I.~Gilitschenski, and D.~Rus, ``Autonomous navigation in inclement weather based on a localizing ground penetrating radar,'' \emph{IEEE Robotics and Automation Letters}, vol.~5, no.~2, pp. 3267--3274, 2020.

\bibitem{Courcelle2023}
C.~Courcelle, D.~Baril, F.~Pomerleau, and J.~Laconte, ``On the importance of quantifying visibility for autonomous vehicles under extreme precipitation,'' \emph{Towards Human-Vehicle Harmonization}, vol.~3, p. 239, 2023.

\bibitem{Churchill2013}
\BIBentryALTinterwordspacing
W.~Churchill and P.~Newman, ``Experience-based navigation for long-term localisation,'' \emph{The International Journal of Robotics Research}, vol.~32, no.~14, pp. 1645--1661, 2013. [Online]. Available: \url{https://doi.org/10.1177/0278364913499193}
\BIBentrySTDinterwordspacing

\bibitem{Gridseth2021}
M.~Gridseth and T.~D. Barfoot, ``Keeping an eye on things: Deep learned features for long-term visual localization,'' \emph{IEEE Robotics and Automation Letters}, vol.~7, no.~2, pp. 1016--1023, 2022.

\bibitem{overbye2023radar}
T.~Overbye and S.~Saripalli, ``Radar-only off-road local navigation,'' \emph{arXiv preprint arXiv:2310.17620}, 2023.

\bibitem{overbye2022g}
------, ``{G-VOM: A GPU accelerated voxel off-road mapping system},'' in \emph{IEEE Intelligent Vehicles Symposium}, 2022, pp. 1480--1486.

\bibitem{michalczyk2023radar}
J.~Michalczyk, M.~Scheiber, R.~Jung, and S.~Weiss, ``Radar-inertial odometry for closed-loop control of resource-constrained aerial platforms,'' in \emph{IEEE International Symposium on Safety, Security, and Rescue Robotics}, 2023, pp. 61--68.

\bibitem{hong2020radarslam}
Z.~Hong, Y.~Petillot, and S.~Wang, ``Radarslam: Radar based large-scale slam in all weathers,'' in \emph{IEEE/RSJ International Conference on Intelligent Robots and Systems}, 2020, pp. 5164--5170.

\bibitem{Adolfsson2021}
D.~Adolfsson, M.~Magnusson, A.~Alhashimi, A.~J. Lilienthal, and H.~Andreasson, ``Oriented surface points for efficient and accurate radar odometry,'' \emph{arXiv preprint arXiv:2109.09994}, 2021.

\bibitem{adolfsson2021cfear}
------, ``{CFEAR} radarodometry-conservative filtering for efficient and accurate radar odometry,'' in \emph{IEEE/RSJ International Conference on Intelligent Robots and Systems}, 2021, pp. 5462--5469.

\bibitem{Adolfsson2022}
------, ``Lidar-level localization with radar? the {CFEAR} approach to accurate, fast, and robust large-scale radar odometry in diverse environments,'' \emph{IEEE Transactions on Robotics}, vol.~39, no.~2, pp. 1476--1495, 2022.

\bibitem{frosi2023advancements}
M.~Frosi, M.~Usuelli, and M.~Matteucci, ``Advancements in radar odometry,'' \emph{arXiv preprint arXiv:2310.12729}, 2023.

\bibitem{alhashimi2021bfar}
A.~Alhashimi, D.~Adolfsson, M.~Magnusson, H.~Andreasson, and A.~J. Lilienthal, ``{BFAR}-bounded false alarm rate detector for improved radar odometry estimation,'' \emph{arXiv preprint arXiv:2109.09669}, 2021.

\bibitem{anderson2015full}
S.~Anderson and T.~D. Barfoot, ``{Full STEAM ahead: Exactly sparse Gaussian process regression for batch continuous-time trajectory estimation on SE (3)},'' in \emph{IEEE/RSJ International Conference on Intelligent Robots and Systems (IROS)}, 2015, pp. 157--164.

\bibitem{Forster2015}
C.~Forster, L.~Carlone, F.~Dellaert, and D.~Scaramuzza, ``Imu preintegration on manifold for efficient visual-inertial maximum-a-posteriori estimation,'' in \emph{Robotics: Science and Systems XI}, 2015.

\bibitem{Rawlings2022}
J.~Rawlings, D.~Mayne, and M.~Diehl, \emph{Model Predictive Control: Theory, Computation, and Design}, 2nd~ed.\hskip 1em plus 0.5em minus 0.4em\relax Nob Hill Publishing.

\bibitem{Andersson2019}
J.~A.~E. Andersson, J.~Gillis, G.~Horn, J.~B. Rawlings, and M.~Diehl, ``{CasADi} -- {A} software framework for nonlinear optimization and optimal control,'' \emph{Mathematical Programming Computation}, vol.~11, no.~1, pp. 1--36, 2019.

\bibitem{Warthog}
\BIBentryALTinterwordspacing
{Clearpath Robotics}. (2020) {Warthog Unmanned Ground Vehicle}. [Online]. Available: \url{https://clearpathrobotics.com/%20warthog-unmanned-ground-vehicle-robot/}
\BIBentrySTDinterwordspacing

\bibitem{ouster_128}
``Ouster {OS1} {LiDAR},'' Available Online [\url{https://ouster.com/products/scanning-lidar/os1-sensor}].

\bibitem{novatelGPS}
``{NovAtel SMART6 GPS},'' Available Online [\url{https://novatel.com/support/previous-generation-products-drop-down/previous-generation-products/smart6-smart-antenna}].

\bibitem{Lisus2024}
D.~Lisus, K.~Burnett, D.~J. Yoon, and T.~D. Barfoot, ``Are doppler velocity measurements useful for spinning radar odometry?'' \emph{arXiv preprint arXiv:2404.01537}, 2024.

\end{thebibliography}
